\documentclass[review]{elsarticle}
\graphicspath{ {./figures/} }
\usepackage{hyperref}
\usepackage{float}
\usepackage{amsmath} 
\usepackage{verbatim} 
\usepackage{apalike}
\restylefloat{figure}
\floatstyle{plaintop} 
\restylefloat{table}
\usepackage{caption}   

\journal{Expert Systems with Applications}

\biboptions{authoryear}

\begin{document}
\begin{frontmatter}


\begin{titlepage}
\begin{center}
\vspace*{1cm}

\textbf{ \large PINT: Physics-Informed Neural Time Series Models with Applications to Long-term Inference on WeatherBench 2m-Temperature Data}

\vspace{1.5cm}

Keonvin Park$^a$ (kbpark16@snu.ac.kr), Jisu Kim$^{a,b,*}$ (jkim82133@snu.ac.kr), Jaemin Seo$^{c,*}$ (jseo@cau.ac.kr) \\
\hspace{10pt}

\begin{flushleft}
\small  
$^a$ Interdisciplinary Program in Artificial Intelligence, Seoul National University, Seoul, Republic of Korea \\
$^b$ Department of Statistics, Seoul National University, Seoul, Republic of Korea \\
$^c$ Department of Physics, Chung-Ang University, Seoul, Republic of Korea

\vspace{1cm}
\textbf{Corresponding authors:} \\
Jisu Kim \\
Department of Statistics, Seoul National University, Seoul, Republic of Korea \\
Tel: +82 2-880-6551 \\
Email: jkim82133@snu.ac.kr \\

Jaemin Seo \\
Department of Physics, Chung-Ang University, Seoul, Republic of Korea \\
Tel: +82 2-820-5169 \\
Email: jseo@cau.ac.kr \\

\end{flushleft}        
\end{center}
\end{titlepage}

\title{PINT: Physics-Informed Neural Time Series Models with Applications to Long-term Inference on WeatherBench 2m-Temperature Data}

\author[label1]{Keonvin Park}
\ead{kbpark16@snu.ac.kr}

\author[label1,label2]{Jisu Kim\corref{cor1}}
\ead{jkim82133@snu.ac.kr}

\author[label3]{Jaemin Seo\corref{cor1}}
\ead{jseo@cau.ac.kr}

\cortext[cor1]{Co-corresponding authors.}
\address[label1]{Interdisciplinary Program in Artificial Intelligence, Seoul National University, Seoul, Republic of Korea}
\address[label2]{Department of Statistics, Seoul National University, Seoul, Republic of Korea}
\address[label3]{Department of Physics, Chung-Ang University, Seoul, Republic of Korea}

\begin{abstract}
This paper introduces \textbf{PINT} (\textit{Physics-Informed Neural Time Series Models}), a novel framework designed to integrate physical constraints into neural time series models, thereby enhancing their ability to capture complex dynamics in real-world datasets. To demonstrate its practical utility, we apply \textbf{PINT} to the ERA5 WeatherBench dataset, a widely-used benchmark for climate prediction, focusing on long-term forecasting of 2m-temperature data. 

\textbf{PINT} leverages the \textit{Simple Harmonic Oscillator Equation} as a physics-informed prior, incorporating its periodic dynamics into three popular neural architectures: RNN, LSTM, and GRU. The choice of the Simple Harmonic Oscillator Equation is motivated by its well-known analytical solutions (sine and cosine functions), which not only represent periodic dynamics but also enable rigorous evaluation of the performance improvements achieved through the incorporation of physics-informed constraints. By benchmarking against a linear regression baseline derived from the exact solutions of this equation, we quantify the added value of embedding physical principles in data-driven models.

Unlike traditional time series approaches that often rely on future observations for inference or training, \textbf{PINT} is designed for practical forecasting scenarios. Using only the first 90 days of observed data, the framework iteratively predicts the next two years, addressing challenges associated with limited or missing real-time updates.

Extensive experiments on the WeatherBench dataset showcase \textbf{PINT}’s ability to generalize to unseen data, accurately capture periodic trends, and align with underlying physical principles. This study highlights the potential of physics-informed neural time series models to bridge the gap between data-driven machine learning and the interpretability required for climate applications.
The implementation of our models and all the datasets used in this study are publicly available on GitHub: https://github.com/KV-Park

\end{abstract}

\end{frontmatter}

\section{Introduction}
\label{introduction}

Accurately modeling periodic dynamics in temporal data is a critical challenge in scientific modeling, with applications spanning physics, biology, and climate science. Real-world processes such as oscillatory systems, seasonal climate variations, and energy cycles exhibit strong periodic behaviors that demand models capable of leveraging domain-specific knowledge for reliable and interpretable predictions. While traditional machine learning approaches excel in short-term forecasting, their ability to generalize to long-term trends while adhering to underlying physical principles remains limited.

This paper introduces \textbf{PINT (Physics-Informed Neural Time Series Models)}, a framework designed to integrate physical constraints into neural architectures for robust handling of periodic patterns in temporal data. Central to the framework is the \textit{Simple Harmonic Oscillator Equation}, a fundamental model of periodic dynamics with well-known analytical solutions (sine and cosine functions). By embedding these solutions as constraints, \textbf{PINT} enhances the ability of neural models to capture oscillatory behaviors and improves their interpretability. Furthermore, the availability of analytical solutions allows us to quantitatively benchmark the performance improvements achieved by incorporating physics-informed constraints.

To evaluate its practical applicability, we apply \textbf{PINT} to the ERA5 WeatherBench dataset \cite{rasp2020weatherbench}, focusing on long-term forecasting of 2m-temperature data. Climate datasets, such as ERA5, are inherently periodic due to Earth's rotation, revolution, and seasonal cycles, making them ideal testbeds for physics-informed approaches. Unlike traditional numerical weather prediction (NWP) models, such as the Integrated Forecasting System (IFS) \cite{bauer2021nwp}, which rely on solving complex partial differential equations (PDEs) for atmospheric simulations, \textbf{PINT} provides a scalable data-driven alternative with reduced computational overhead and improved interpretability.

In comparison to existing machine learning approaches like ClimODE \cite{verma2024climode}, ClimaX \cite{nguyen2023climax}, and FourCastNet \cite{pathak2022fourcastnet}, which focus on accurate short-term numerical predictions,  \textbf{PINT} differentiates itself by focusing on long-term forecasting. By iteratively predicting extended future trends using only initial observations, PINT addresses challenges associated with cumulative errors and trend fidelity in long-term inference. Furthermore, by benchmarking against a linear regression baseline derived from the harmonic oscillator's analytical solutions, we ensure rigorous and interpretable evaluation of the proposed framework’s performance.

\textbf{Key Contributions:}
\begin{itemize}
    \item We introduce \textbf{PINT}, a Physics-Informed Neural Time Series framework, integrating the \textit{Simple Harmonic Oscillator Equation} as a constraint to capture periodic dynamics in temporal data.
    \item We demonstrate \textbf{PINT}’s applicability on the ERA5 WeatherBench dataset for long-term forecasting, showcasing its ability to model periodic trends inherent in real-world climate data.
    \item By benchmarking against a baseline linear regression model based on the harmonic oscillator’s exact solutions, we quantitatively assess the benefits of embedding physical constraints in neural architectures.
        \item Unlike traditional RNN, LSTM, and GRU architectures, which also can perform auto-regressive inference, \textbf{PINT} leverages embedded physical laws to deliver more robust and interpretable results for long-term forecasting.
\end{itemize}

\section{Methodology}
\label{methods}
\subsection{Recurrent Neural Networks (RNN)}
Recurrent Neural Networks (RNN) are a foundational class of deep learning models designed to handle sequential data. By employing recurrent connections, RNN can capture temporal dependencies in data, making them suitable for time-series forecasting tasks. However, traditional RNN often suffer from the vanishing gradient problem, which hampers their ability to learn long-term dependencies \cite{hopfield1982rnn}. Despite this limitation, RNN provide a baseline for evaluating the effectiveness of more advanced architectures.

The RNN architecture processes input sequences iteratively. At each time step \( t \), the hidden state \( h_t \) is updated using the input \( x_t \) and the hidden state from the previous time step \( h_{t-1} \). This iterative computation enables the network to learn temporal patterns and dependencies in sequential data. The recursive nature of RNN allows it to maintain information over time, but the effectiveness of this mechanism diminishes as the sequence length increases, leading to challenges such as forgetting long-term dependencies.

\subsection{Long Short-Term Memory Networks (LSTM)}
Long Short-Term Memory Networks (LSTM) extend RNNs by introducing memory cells and gating mechanisms—namely, the input, forget, and output gates—that regulate the flow of information \cite{hochreiter1997lstm}. These gates enable LSTM to learn and retain long-term dependencies effectively, overcoming the vanishing gradient problem. Due to their ability to model complex temporal relationships, LSTM are widely used in climate forecasting tasks \cite{qin2017dual}.

\subsection{Gated Recurrent Units (GRU)}
Gated Recurrent Units (GRU) simplify LSTM by merging the input and forget gates into a single update gate \cite{cho2014gru}. This reduction in complexity decreases the number of parameters while maintaining the ability to model long-term dependencies. GRU have shown comparable performance to LSTM in various sequential modeling tasks, including weather prediction \cite{yu2017spatiotemporal}.

\subsection{Physics-Informed Neural Time Series Models (PINT)}
Physics-informed neural networks (PINN) incorporate domain knowledge by embedding physical constraints into the learning process \cite{raissi2019pinns}. For example, in climate modeling, seasonal variations can be described by the harmonic oscillator equation:
\begin{equation}
    u''(t) + \omega^2 u(t) = 0,
\end{equation}
where \( \omega = \frac{2\pi}{T} \) represents the angular frequency for a periodic signal with period \( T \). Given that the period \( T \) for climate data corresponds to one year (365 days), \( \omega \) is set to \( \frac{2\pi}{365} \).

\begin{figure}[h!]
    \centering
    \includegraphics[width=0.9\textwidth]{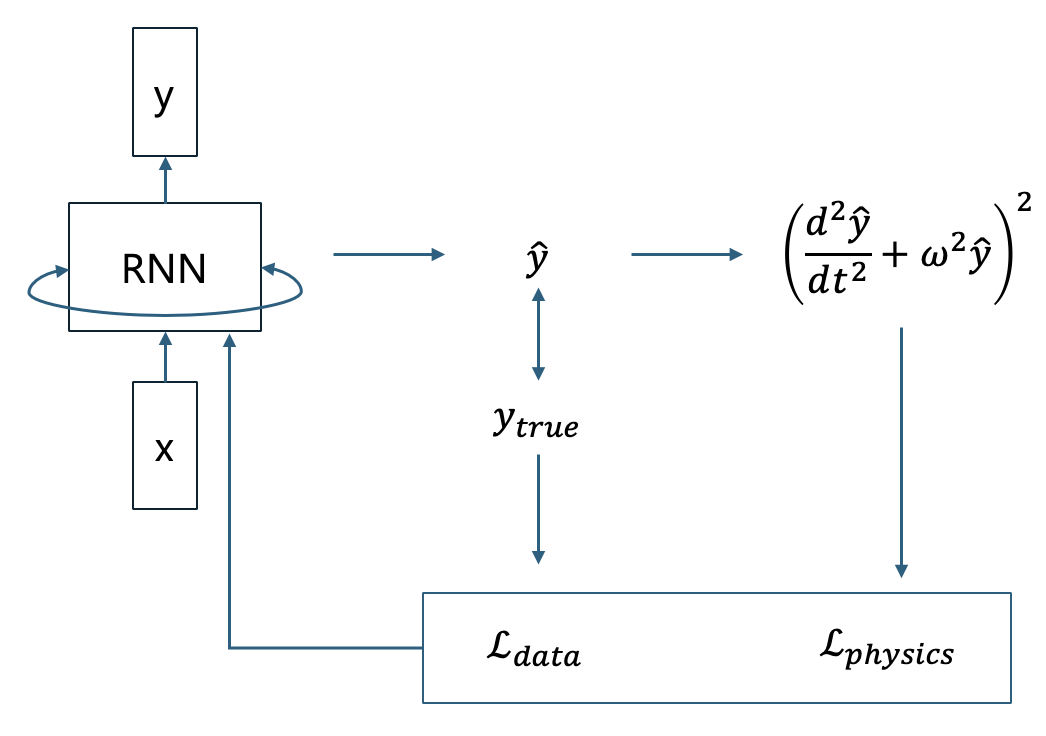}
    \caption{Structure of PINT utilizing physics knowledge based on Simple harmonic oscillator.}
    \label{fig:rnn_structure}
\end{figure}

Notably, since the data has been standardized (mean centered and variance normalized), no offset value (constant term) is required. The standardization ensures that the mean of the data is zero, effectively aligning the harmonic oscillator equation with the standardized time series.
Moreover, as shown in Figure~\ref{fig:rnn_structure}, the framework combines data-driven loss with physics-based constraints to enhance model predictions.

The corresponding physics-informed loss ensures that the model predictions align with known physical laws:
\begin{equation}
    \mathcal{L}_{\text{physics}} = \frac{1}{N} \sum_{i=1}^N \left( u''(t_i) + \omega^2 u(t_i) \right)^2.
\end{equation}

The total loss function used during training combines the physics-informed loss with the standard data-driven loss, weighted by a hyperparameter \( \lambda_{\text{physics}} \), which balances the influence of the two terms:
\begin{equation}
    \mathcal{L}_{\text{total}} = \lambda_{\text{data}}\mathcal{L}_{\text{data}} + \lambda_{\text{physics}} \mathcal{L}_{\text{physics}},
\end{equation}
where \( \mathcal{L}_{\text{data}} \) represents the standard data loss (e.g., mean squared error between predictions and observations), \( \lambda_{\text{data}} \) is the data loss weight which is set to 1, and \( \lambda_{\text{physics}} \) is the physics loss weight. This combined loss formulation enables the model to leverage prior knowledge effectively while ensuring accurate predictions from the data.

Integrating this combined loss term into the training process improves interpretability and generalization, especially for tasks requiring the enforcement of physical consistency, such as climate forecasting.

\subsection{Linear Regression under Physics's Law}
Linear Regression serves as a fundamental statistical technique used to model the relationship between a dependent variable and one or more independent variables by fitting a linear equation to observed data. In the context of climate modeling, a simple harmonic equation is often sufficient for representing seasonal cycles:

\begin{equation}
    x(t) = \beta_1 \cos(\omega t) + \beta_2 \sin(\omega t),
\end{equation}

where \( \omega = \frac{2\pi}{T} \). For this study, the period \( T \) is set to 365 days, resulting in \( \omega = \frac{2\pi}{365} \). This formulation mirrors the basic structure of a simple harmonic oscillator, providing a straightforward yet effective baseline for comparing more sophisticated machine learning approaches.

Notably, since the dependent variable \( x(t) \) has been standardized (mean centered and variance normalized), no offset value (constant term) is required in this formulation. The standardization ensures that the mean of the data is zero, thereby eliminating the need for a constant intercept term.

\subsection{Autoregressive Inference for Long-Term Forecasting}

In the context of climate modeling, autoregressive inference is a practical approach for long-term forecasting. Unlike standard predictive models that rely on complete input sequences for each prediction step, autoregressive models iteratively use their own predictions as inputs for subsequent forecasts. This allows them to extend forecasting horizons without requiring intermediate observations.

The autoregressive process begins by using an initial sequence of observations as input to the model. The model generates a set of predictions for a specified time horizon, which are then concatenated with the original input sequence to form a new input. This process is repeated iteratively to achieve long-term forecasts, leveraging the model’s ability to capture both short-term dynamics and long-term trends.

\begin{figure}[h!]
    \centering
    \includegraphics[width=0.96\textwidth]{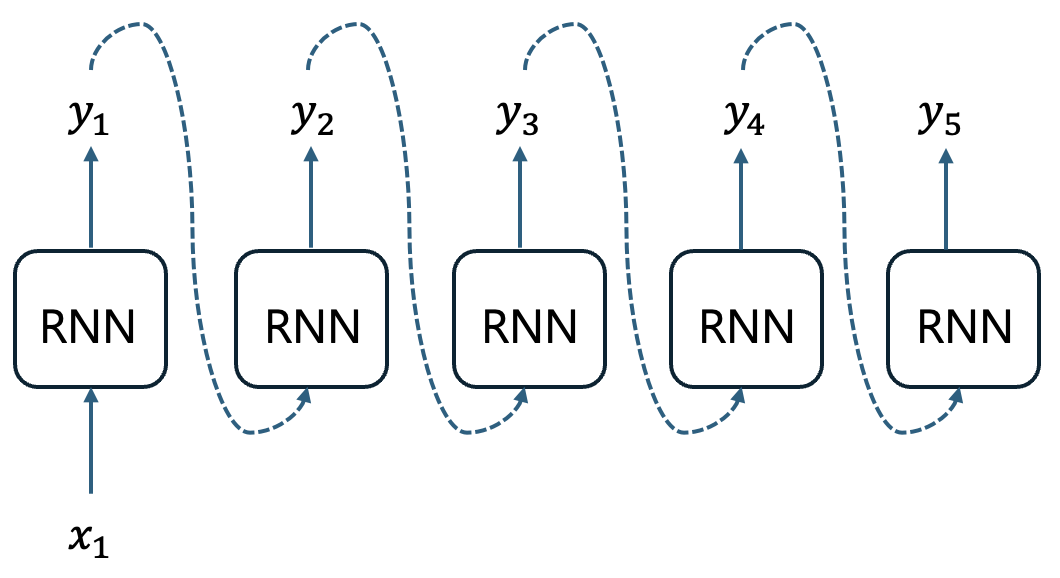}
    \caption{Workflow of the autoregressive inference for long-term forecasting.}
    \label{fig:autoregressive_structure}
\end{figure}

Figure~\ref{fig:autoregressive_structure} illustrates the autoregressive inference workflow in the context of Recurrent Neural Networks (RNN). In each step:
\begin{itemize}
    \item The model takes an input vector \( x \), which consists of observed data for the initial time steps.
    \item The model predicts the next set of values \( \hat{y} \) for the forecasting horizon.
    \item The predicted values \( \hat{y} \) are concatenated with the original input \( x \), forming a new input sequence for the next iteration.
\end{itemize}

This iterative process enables autoregressive models to forecast for extended periods while maintaining consistency with previously predicted values, making them particularly suitable for long-term climate predictions.

\subsubsection{Performance Metrics}
To evaluate the models, two key performance metrics are used:
\begin{itemize}
    \item \textbf{Root Mean Square Error (RMSE)}: This measures the average magnitude of the prediction errors, defined as:
    \begin{equation}
        \text{RMSE} = \sqrt{\frac{1}{N} \sum_{i=1}^N (y_i - \hat{y}_i)^2},
    \end{equation}
    where \( y_i \) are the observed values and \( \hat{y}_i \) are the predicted values. A lower RMSE indicates better model accuracy.

    \item \textbf{Correlation Coefficient (CORR)}: This quantifies the strength and direction of the linear relationship between the observed and predicted values, defined as:
    \begin{equation}
        \text{CORR} = \frac{\sum_{i=1}^N (y_i - \bar{y})(\hat{y}_i - \bar{\hat{y}})}{\sqrt{\sum_{i=1}^N (y_i - \bar{y})^2 \sum_{i=1}^N (\hat{y}_i - \bar{\hat{y}})^2}},
    \end{equation}
    where \( \bar{y} \) and \( \bar{\hat{y}} \) are the means of the observed and predicted values, respectively. A CORR value closer to 1 indicates better alignment between predictions and observations.
\end{itemize}
\section{Data}

To evaluate the proposed framework, we utilize the ERA5 dataset from WeatherBench, a comprehensive real-world time series dataset specifically designed for benchmarking weather forecasting models.

\subsection{ERA5 WeatherBench}
The ERA5 dataset, derived from the European Centre for Medium-Range Weather Forecasts (ECMWF), offers a global atmospheric reanalysis that has been widely utilized in climatic studies \cite{rasp2020weatherbench}. For this study, we specifically focus on the \textit{2m\_temperature (t2m)} variable from WeatherBench, which records the air temperature at two meters above the ground. The \textit{t2m} variable is particularly suitable for our analysis as it reflects near-surface atmospheric conditions, making it highly relevant for climate modeling and forecasting applications.

The dataset includes hourly measurements for the year 1979 onwards, which we have aggregated into daily means to facilitate our analysis. By leveraging the \textit{t2m} variable, we aim to capture both short-term temperature fluctuations and long-term seasonal trends, which are essential for evaluating the performance of the proposed framework.

We selected three cities with distinct climatic conditions—Seoul (South Korea), Washington, D.C. (United States), and Beijing (China)—to ensure the robustness and applicability of our model across different geographic and environmental contexts. The data for each city is spatially averaged around the nearest grid point to the city's coordinates, ensuring that local climatic conditions are adequately captured.

\subsection{Data Processing and Sequence Modeling}
The temperature data from WeatherBench is split into training, validation, and test sets covering different years. Moreover, descriptive statistics for each is shown in Table~\ref{tab:realworld_data_overview}:
\begin{itemize}
    \item \textbf{Training Data:} 2008-2012
    \item \textbf{Validation Data:} 2013-2015
    \item \textbf{Test Data:} 2016-2018
\end{itemize}

Each dataset includes daily temperature averages, which are used to construct input sequences of 90 days. These sequences are utilized to predict the subsequent 30 days of temperatures, allowing the model to capture both short-term fluctuations and seasonal trends. This sequence modeling approach is designed to test the model's ability to leverage past climate data to forecast future conditions effectively.

\begin{table}[h!]
\centering
\caption{\normalsize Descriptive Statistics for 2m-temperature (t2m)}
\begin{tabular}{|c|c|c|c|}
\hline
\textbf{City} & \textbf{Dataset} & \textbf{Mean (°C)} & \textbf{Std (°C)} \\ \hline
              & Train (2008-2012) & 12.96 & 9.06 \\ 
Seoul         & Validation (2013-2015) & 13.43 & 8.76 \\ 
              & Test (2016-2018) & 13.58 & 9.07 \\ \hline
              & Train (2008-2012) & 4.49 & 13.16 \\ 
Beijing       & Validation (2013-2015) & 5.16 & 12.26 \\ 
              & Test (2016-2018) & 5.34 & 12.93 \\ \hline
              & Train (2008-2012) & 18.68 & 4.19 \\ 
Washington-DC & Validation (2013-2015) & 18.90 & 4.20 \\ 
              & Test (2016-2018) & 19.03 & 4.10 \\ \hline
\end{tabular}
\label{tab:realworld_data_overview}
\end{table}
\section{Results}

The predictive performance of the models was evaluated using two key metrics: Root Mean Square Error (RMSE) and correlation coefficient (CORR). RMSE provides a measure of the average magnitude of errors between predicted and observed temperatures, offering a direct comparison of model accuracy. CORR evaluates the strength and direction of the linear relationship between predictions and actual values, reflecting the models' ability to capture temporal trends in the data.
Figures~\ref{fig:physics_timeseries_prediction}, \ref{fig:overall_timeseries_prediction} illustrate the time series comparisons for Seoul, Beijing, and Washington-DC, respectively, highlighting the performance of different models across these cities. Moreover, hyperparameters for RNN-family models and PINN-based models are presented in Table~\ref{tab:rnn_hyperparameters}, Table~\ref{tab:pinn_hyperparameters}, respectively. Furthermore, Estimated Linear Regression Beta Coefficients for each city are presented in Table~\ref{tab:linear_regression_results}.

\begin{table}[h!]
\centering
\caption{Hyperparameters configuration of the RNN family used.}
\label{tab:rnn_hyperparameters}
\begin{tabular}{|l|l|}
\hline
\textbf{Hyperparameter}      & \textbf{Configuration} \\ \hline
Optimizer                   & Adam                   \\ \hline
Weight initialization       & Random                 \\ \hline
Learning rate               & 0.001                  \\ \hline
Activation function         & tanh                   \\ \hline
Dropout rate                & 0.1                    \\ \hline
Number of hidden layers     & 2                      \\ \hline
Number of neurons per layer & 64                     \\ \hline
Sequence length             & 90                     \\ \hline
Prediction length           & 30                     \\ \hline
Training epochs             & 1000                  \\ \hline
Batch size                  & Full batch             \\ \hline
Early stopping              & No                    \\ \hline
\end{tabular}
\end{table}

\begin{table}[h!]
\centering
\caption{Hyperparameters configuration of the PINN used.}
\label{tab:pinn_hyperparameters}
\begin{tabular}{|l|l|}
\hline
\textbf{Hyperparameter}      & \textbf{Configuration} \\ \hline
Optimizer                   & Adam                   \\ \hline
Weight initialization       & Random                 \\ \hline
Learning rate               & 0.001                  \\ \hline
Activation function         & tanh                   \\ \hline
Data loss weight            & 1                      \\ \hline
Physics loss weight         & 0.001                  \\ \hline
Regularization weight       & -                      \\ \hline
Number of hidden layers     & 2                      \\ \hline
Number of neurons per layer & 64                     \\ \hline
Dropout rate                & 0.1                    \\ \hline
Training epochs             & 1000                  \\ \hline
Early stopping              & No                    \\ \hline
Batch size                  & Full batch             \\ \hline
\end{tabular}
\end{table}

\begin{table}[h!]
\centering
\caption{Linear Regression Beta Coefficients}
\label{tab:linear_regression_results}
\begin{tabular}{|c|c|c|}
\hline
\textbf{City} & \(\hat{\beta}_1\)(cosine) & \(\hat{\beta}_2\)(sine)  \\ \hline
Seoul         & -1.2038     & -0.5845      \\ \hline
Beijing       & -1.3088     & -0.3113      \\ \hline
Washington-DC & -1.0197     & -0.8799      \\ \hline
\end{tabular}
\end{table}
This section evaluates the performance of different models, focusing on three key comparisons: (1) RNN-family models vs. Physics-Informed Counterparts, (2) A Best Model vs. Linear Regression as a baseline, and (3) general observations across datasets.

\subsection{Comparison 1: RNN-family Models vs. Physics-Informed Counterparts}

\begin{figure}[h!]
    \centering
    \includegraphics[width=0.96\textwidth]{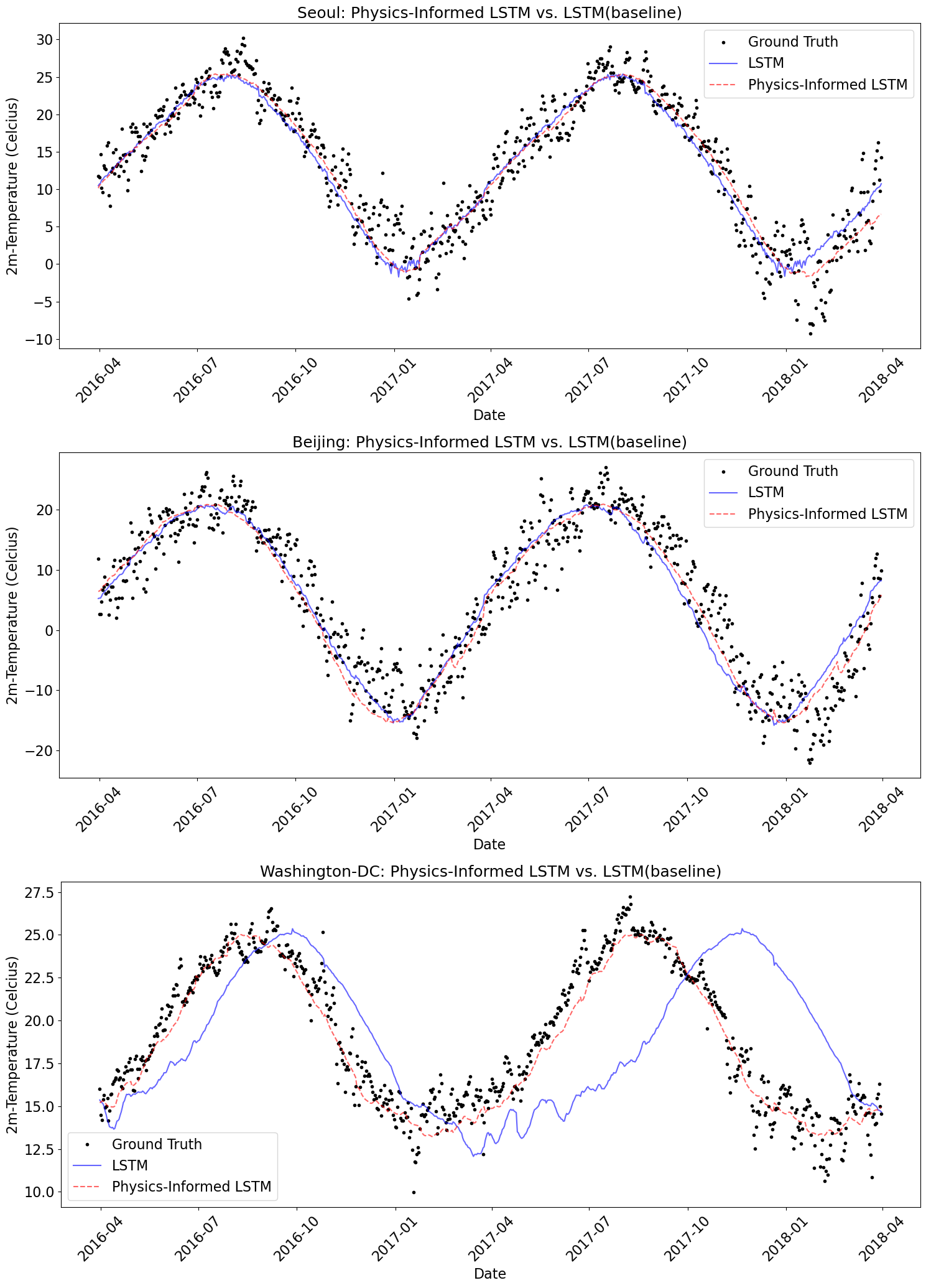}
    \caption{A Comparison for RNN-family Models vs. Physics-Informed Counterparts across cities.}
    \label{fig:physics_timeseries_prediction}
\end{figure}

\begin{figure}[h!]
    \centering
    \includegraphics[width=0.8\textwidth]{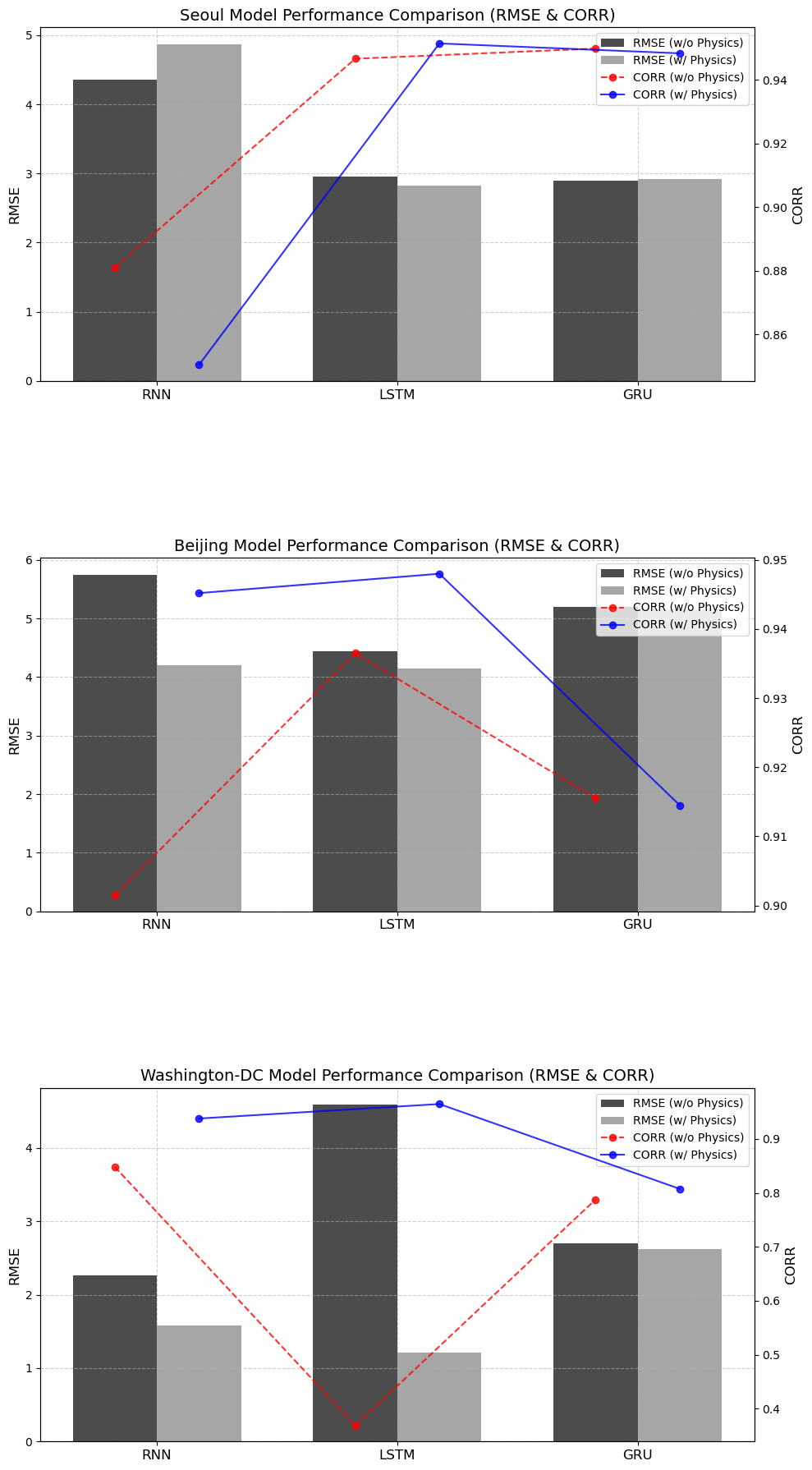}
    \caption{A Comparison for RNN-family Models vs. Physics-Informed Counterparts across cities.}
    \label{fig:Physics_Comparison}
\end{figure}

Physics-Informed Neural Networks (PINNs) were developed to enhance the accuracy and interpretability of RNN-family models by embedding physical constraints inspired by the harmonic oscillator equation. Among the tested models, the Physics-Informed LSTM consistently delivered the best performance across the three cities analyzed, underscoring the value of incorporating physical principles into machine learning models. Results are summarized in Table~\ref{tab:seoul_performance}, Table~\ref{tab:beijing_performance}, and Table~\ref{tab:washington_performance}. A comprehensive visual comparison is presented in Figure~\ref{fig:physics_timeseries_prediction} and Figure~\ref{fig:Physics_Comparison}.

\begin{table}[h!]
\centering
\caption{Seoul Model Performance Summary}
\label{tab:seoul_performance}
\begin{tabular}{|l|l|l|l|l|}
\hline
Model & RMSE & RMSE (Physics) & CORR & CORR (Physics) \\ \hline
RNN   & 4.3558         & 4.8702 (+0.5144) & 0.8811         & 0.8505 (-0.0306) \\ \hline
LSTM  & 2.9612         & \textbf{2.8280 (-0.1332)} & 0.9467         & \textbf{0.9515 (+0.0048)} \\ \hline
GRU   & 2.8939         & 2.9151 (+0.0212) & 0.9499         & 0.9484 (-0.0015) \\ \hline
\end{tabular}
\end{table}

\begin{table}[h!]
\centering
\caption{Beijing Model Performance Summary}
\label{tab:beijing_performance}
\begin{tabular}{|l|l|l|l|l|}
\hline
Model & RMSE & RMSE (Physics) & CORR & CORR (Physics) \\ \hline
RNN   & 5.7525         & \textbf{4.1982 (-1.5543)} & 0.9015         & \textbf{0.9452 (+0.0437)} \\ \hline
LSTM  & 4.4475         & \textbf{4.1492 (-0.2981)} & 0.9365         & \textbf{0.9480 (+0.0115)} \\ \hline
GRU   & 5.1939         & \textbf{5.1521 (-0.0418)} & 0.9155         & \textbf{0.9145 (-0.0010)} \\ \hline
\end{tabular}
\end{table}

\begin{table}[h!]
\centering
\caption{Washington-DC Model Performance Summary}
\label{tab:washington_performance}
\begin{tabular}{|l|l|l|l|l|}
\hline
Model & RMSE & RMSE (Physics) & CORR & CORR (Physics) \\ \hline
RNN   & 2.2672          & \textbf{1.5787 (-0.6885)} & 0.8482          & \textbf{0.9380 (+0.0898)} \\ \hline
LSTM  & 4.5893          & \textbf{1.2076 (-3.3817)} & 0.3687          & \textbf{0.9649 (+0.5962)} \\ \hline
GRU   & 2.7017          & \textbf{2.6218 (-0.0799)} & 0.7879          & \textbf{0.8075 (+0.0196)} \\ \hline
\end{tabular}
\end{table}

\begin{itemize}
    \item \textbf{Seoul: Physics-Informed LSTM Outperforms All Models}  
        In Seoul, which exhibited strong seasonal trends, the Physics-Informed LSTM achieved the best performance:
        \begin{itemize}
            \item \textbf{CORR:} 0.9515
            \item \textbf{RMSE:} 2.8280
        \end{itemize}
        The Physics-Informed GRU performed similarly but with slightly higher RMSE, emphasizing the effectiveness of LSTM-based architectures for periodic datasets.

    \item \textbf{Beijing: Physics-Informed LSTM Maintains Superiority}  
        In Beijing, the Physics-Informed LSTM once again delivered the best balance between RMSE and CORR:
        \begin{itemize}
            \item \textbf{CORR:} 0.9480
            \item \textbf{RMSE:} 4.1492
        \end{itemize}
        The Physics-Informed LSTM demonstrated superior performance across all metrics, highlighting its ability to effectively capture temporal patterns and trends in the data.

    \item \textbf{Washington-DC: Physics-Informed LSTM Excels Despite Stable Trends}  
        In Washington-DC, the Physics-Informed LSTM significantly outperformed all other models:
        \begin{itemize}
            \item \textbf{CORR:} 0.9649
            \item \textbf{RMSE:} 1.2076
        \end{itemize}
        These results highlight the LSTM's ability to model complex dependencies and the added benefit of embedding physical constraints, even in datasets with relatively stable temporal variations.
\end{itemize}

\subsection{Comparison 2: A Best Model vs. Linear Regression (Baseline)}

\begin{figure}[h!]
    \centering
    \includegraphics[width=0.96\textwidth]{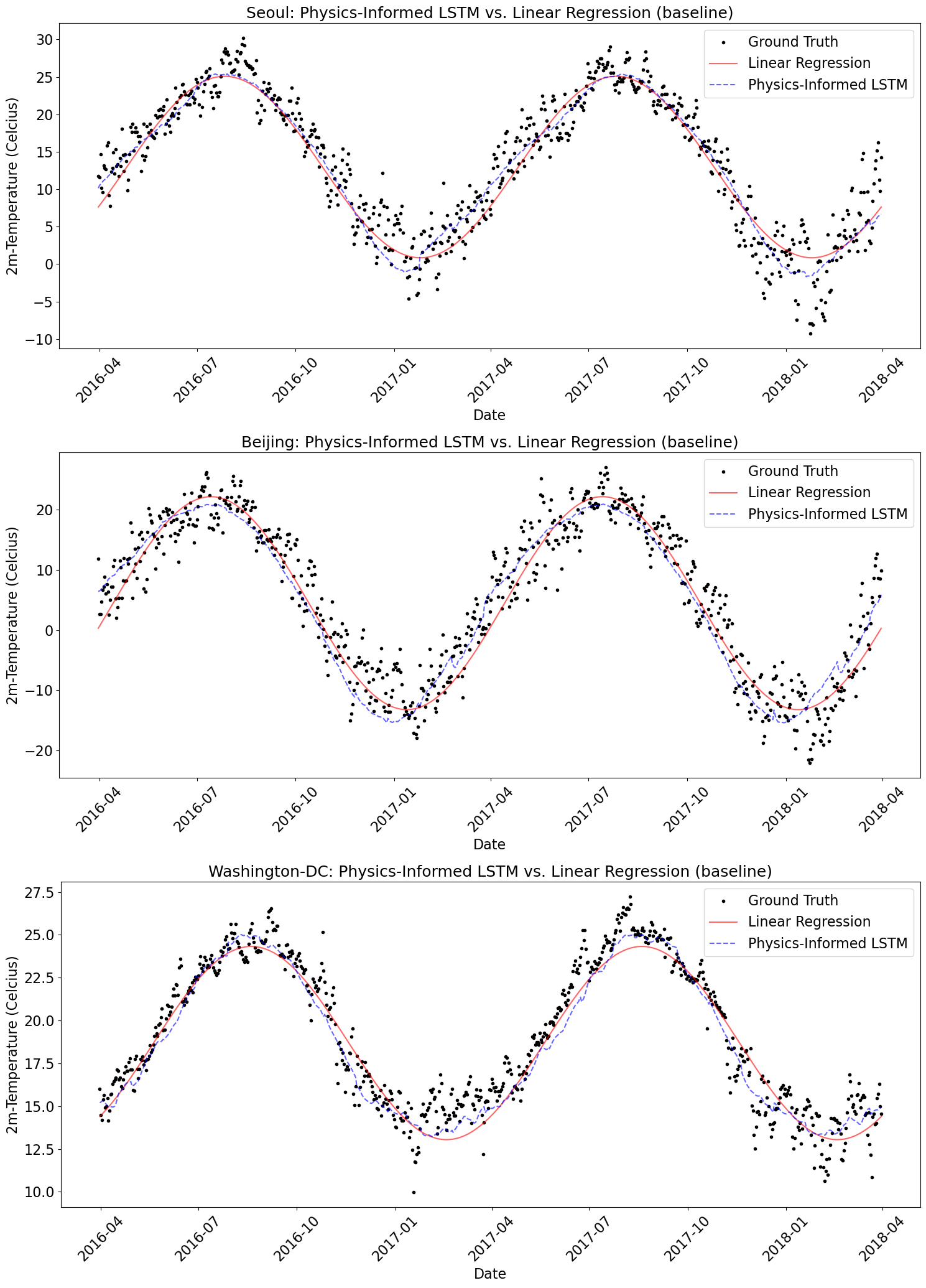}
    \caption{A Comparison for Best Model per city vs. Linear Regression (baseline).}
    \label{fig:overall_timeseries_prediction}
\end{figure}

\begin{figure}[h!]
    \centering
    \includegraphics[width=0.96\textwidth]{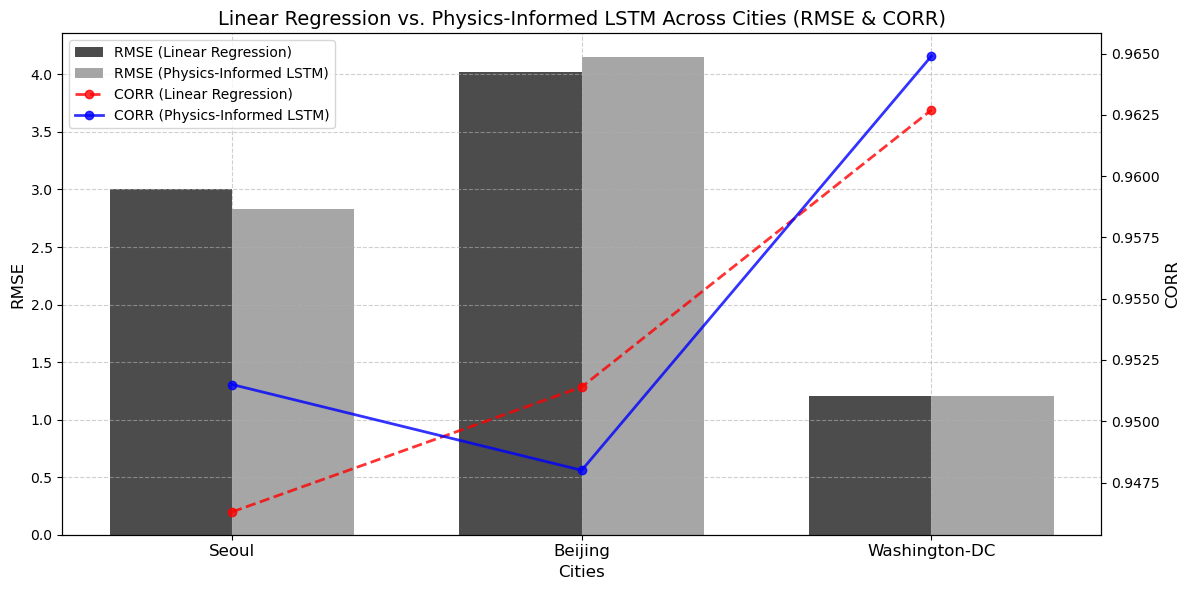}
    \caption{A Comparison for Best Model per city vs. Linear Regression (baseline).}
    \label{fig:overall_figure}
\end{figure}

Linear Regression served as a baseline model, leveraging harmonic oscillator-based seasonal trends for interpretable predictions. Unlike deep learning models, Linear Regression required only the training set (2008–2012) and avoided the need for a validation set. Despite its simplicity, it performed competitively in Beijing and Washington-DC, highlighting the effectiveness of simpler models in certain contexts. For visual comparisons, see Figure~\ref{fig:overall_timeseries_prediction} and Figure~\ref{fig:overall_figure}.

\begin{table}[h!]
\centering
\caption{A Comparison of the Best Model and Linear Regression (2016–2018)}
\label{tab:comparison_best_lr}
\begin{tabular}{|c|c|c|c|c|}
\hline
\textbf{City} & \textbf{Year} & \textbf{Model} & \textbf{RMSE} & \textbf{CORR} \\ \hline
Seoul         & 2016–2018 & Linear Regression       & 3.0015       & 0.9463       \\ 
              &           & Physics-Informed LSTM    & \textbf{2.8280} & \textbf{0.9515} \\ \hline
Beijing       & 2016–2018 & Linear Regression       & \textbf{4.0187}       & \textbf{0.9514} \\ 
              &           & Physics-Informed LSTM    & 4.1492 & 0.9480       \\ \hline
Washington-DC & 2016–2018 & Linear Regression       & 1.2065 & 0.9627 \\ 
              &           & Physics-Informed LSTM                     & \textbf{1.2076}       & \textbf{0.9649}       \\ \hline
\end{tabular}
\end{table}

\begin{itemize}
    \item \textbf{Seoul: Physics-Informed LSTM Outperforms Baseline}  
        The Physics-Informed LSTM significantly outperformed Linear Regression:
        \begin{itemize}
            \item \textbf{RMSE:} 2.8280 (vs. 3.0015 for Linear Regression)
            \item \textbf{CORR:} 0.9515 (vs. 0.9463 for Linear Regression)
        \end{itemize}
        These results underscore the advantage of combining data-driven approaches with physical constraints in datasets with pronounced seasonal trends.

    \item \textbf{Beijing: Linear Regression Remains Competitive}  
        In Beijing, Linear Regression outperformed all models, including the Physics-Informed LSTM:
        \begin{itemize}
            \item \textbf{RMSE:} 4.0187
            \item \textbf{CORR:} 0.9514
        \end{itemize}
        This result highlights the robustness of simpler models in datasets dominated by seasonal patterns. Nevertheless, as shown in Figure~\ref{fig:overall_timeseries_prediction}, the Physics-Informed LSTM better captured finer local trends than Linear Regression.

    \item \textbf{Washington-DC: Physics-Informed LSTM Matches Linear Regression}  
        In Washington-DC, the Physics-Informed LSTM achieved comparable performance to Linear Regression:
        \begin{itemize}
            \item \textbf{RMSE:} 1.2076 (vs. 1.2065 for Linear Regression)
            \item \textbf{CORR:} 0.9649 (vs. 0.9627 for Linear Regression)
        \end{itemize}
        The results demonstrate that the Physics-Informed LSTM effectively captures stable trends while maintaining physical interpretability.
\end{itemize}

\subsection{General Observations}

\begin{itemize}
    \item PINNs excel in capturing periodic trends and long-term temporal dynamics in datasets with pronounced seasonal variability, as observed in Seoul.
    \item Simpler models like Linear Regression remain competitive for datasets with stable seasonal patterns (e.g., Beijing).
    \item For datasets with low variability and stable trends (e.g., Washington-DC), the Physics-Informed LSTM matched the performance of Linear Regression while offering physical interpretability.
\end{itemize}

\section{Conclusion and Future Work}

This study highlights the effectiveness of integrating physics-based constraints into neural network architectures for long-term forecasting tasks. The proposed framework, \textbf{PINT (Physics-Informed Neural Time Series Models)}, has been demonstrated on climate datasets, specifically for predicting near-surface air temperature (\textit{t2m}), showcasing its ability to capture complex temporal dynamics and long-term trends. However, \textbf{PINT} is designed as a \textit{general-purpose physics-informed framework} that can be extended to various domains beyond climate prediction, offering significant potential in fields with underlying periodic or physical dynamics, such as energy forecasting, finance, and healthcare.

The results indicate that \textbf{PINT} delivers superior performance compared to traditional recurrent neural networks (RNNs) for datasets with pronounced seasonal patterns. At the same time, simpler models like Linear Regression remain competitive for datasets with stable, well-defined seasonal trends. These findings underscore the importance of selecting models based on the nature and complexity of the dataset, balancing data-driven learning with physical interpretability.

\subsection*{Future Directions}

To further extend and enhance the utility of \textbf{PINT}, several promising directions for future research are identified:

\begin{itemize}
    \item \textbf{Generalization Across Domains:} While this study focused on climate datasets, future work should explore the application of \textbf{PINT} to other domains where periodicity and physical constraints are prevalent, such as renewable energy production, stock market analysis, and biological systems modeling.
    
    \item \textbf{Multivariate Forecasting:} Expanding the framework to incorporate multivariate time series forecasting could offer richer insights. Variables like humidity, precipitation, and wind speed in climate science—or other interdependent factors in different domains—should be integrated to capture nonlinear interactions and dependencies.

    \item \textbf{Adaptive Physics Loss:} Developing adaptive weighting mechanisms for the physics-informed loss terms will improve the trade-off between physical constraints and data-driven learning. This will enhance the generalizability of \textbf{PINT} across diverse datasets and temporal resolutions.

    \item \textbf{Regional and Global Evaluation:} For climate science, future studies should evaluate \textbf{PINT} across a broader range of geographic regions, including polar, tropical, and arid zones, to validate the model's robustness under varying climatic regimes. Similarly, evaluations in other domains should address diverse use cases to demonstrate the framework's adaptability.

    \item \textbf{Uncertainty Quantification:} Integrating uncertainty quantification techniques into \textbf{PINT} will enable probabilistic predictions, improving reliability and interpretability. This is especially valuable for real-world applications where decision-making depends on understanding prediction confidence.

\end{itemize}

\subsection*{Final Remarks}

This research underscores the potential of \textbf{PINT} in bridging the gap between traditional physics-based models and data-driven approaches for long-term forecasting. By leveraging domain knowledge through physics constraints, \textbf{PINT} offers significant advantages in capturing intricate patterns and long-term trends, while retaining interpretability and robustness. At the same time, the study highlights the importance of benchmarking against simple yet effective models, such as Linear Regression, to ensure a balanced and context-aware approach to predictive modeling.

The general-purpose nature of \textbf{PINT} provides a pathway for broader adoption across various disciplines, fostering advancements in predictive accuracy and applicability. By enabling more actionable and interpretable insights, \textbf{PINT} contributes to the development of models that are both scientifically rigorous and practically impactful, paving the way for transformative innovations in data-driven forecasting and decision-making.

\section*{Declaration of Competing Interest}
The authors declare that they have no known competing financial interests or personal relationships that could have appeared to influence the work reported in this paper.

\section*{Acknowledgements}
This work was partly supported by Institute of Information \& communications Technology Planning \& Evaluation (IITP) grant funded by the Korea government(MSIT) [NO.RS-2021-II211343, Artificial Intelligence Graduate School Program (Seoul National University)], and Chung-Ang University.

\section*{Data availability}
The datasets used for the analysis are publicly available and can be accessed from [https://github.com/pangeo-data/WeatherBench?tab=readme-ov-file].

\bibliography{sample}

\begin{thebibliography}{11}
\expandafter\ifx\csname natexlab\endcsname\relax\def\natexlab#1{#1}\fi
\providecommand{\url}[1]{\texttt{#1}}
\providecommand{\href}[2]{#2}
\providecommand{\path}[1]{#1}
\providecommand{\DOIprefix}{doi:}
\providecommand{\ArXivprefix}{arXiv:}
\providecommand{\URLprefix}{URL: }
\providecommand{\Pubmedprefix}{pmid:}
\providecommand{\doi}[1]{\href{http://dx.doi.org/#1}{\path{#1}}}
\providecommand{\Pubmed}[1]{\href{pmid:#1}{\path{#1}}}
\providecommand{\bibinfo}[2]{#2}
\ifx\xfnm\relax \def\xfnm[#1]{\unskip,\space#1}\fi
\bibitem[{Bauer et~al.(2021)Bauer, Thorpe \& Brunet}]{bauer2021nwp}
\bibinfo{author}{Bauer, P.}, \bibinfo{author}{Thorpe, A.}, \& \bibinfo{author}{Brunet, G.} (\bibinfo{year}{2021}).
\newblock \bibinfo{title}{The quiet revolution of numerical weather prediction}.
\newblock {\it \bibinfo{journal}{Nature}\/},  {\it \bibinfo{volume}{525}\/}, \bibinfo{pages}{47--55}.
\bibitem[{Cho et~al.(2014)Cho, Van~Merriënboer, Gulcehre, Bahdanau, Bougares, Schwenk \& Bengio}]{cho2014gru}
\bibinfo{author}{Cho, K.}, \bibinfo{author}{Van~Merriënboer, B.}, \bibinfo{author}{Gulcehre, C.}, \bibinfo{author}{Bahdanau, D.}, \bibinfo{author}{Bougares, F.}, \bibinfo{author}{Schwenk, H.}, \& \bibinfo{author}{Bengio, Y.} (\bibinfo{year}{2014}).
\newblock \bibinfo{title}{Learning phrase representations using rnn encoder–decoder for statistical machine translation}.
\newblock In {\it \bibinfo{booktitle}{Proceedings of the 2014 Conference on Empirical Methods in Natural Language Processing (EMNLP)}\/} (pp. \bibinfo{pages}{1724--1734}).
\newblock \bibinfo{address}{Doha, Qatar}.
\bibitem[{Hochreiter \& Schmidhuber(1997)}]{hochreiter1997lstm}
\bibinfo{author}{Hochreiter, S.}, \& \bibinfo{author}{Schmidhuber, J.} (\bibinfo{year}{1997}).
\newblock \bibinfo{title}{Long short-term memory}.
\newblock {\it \bibinfo{journal}{Neural Computation}\/},  {\it \bibinfo{volume}{9}\/}, \bibinfo{pages}{1735--1780}.
\bibitem[{Hopfield(1982)}]{hopfield1982rnn}
\bibinfo{author}{Hopfield, J.~J.} (\bibinfo{year}{1982}).
\newblock \bibinfo{title}{Neural networks and physical systems with emergent collective computational abilities}.
\newblock {\it \bibinfo{journal}{Proceedings of the National Academy of Sciences}\/},  {\it \bibinfo{volume}{79}\/}, \bibinfo{pages}{2554--2558}.
\bibitem[{Nguyen et~al.(2023)Nguyen, Hu, He \& et~al.}]{nguyen2023climax}
\bibinfo{author}{Nguyen, H.}, \bibinfo{author}{Hu, K.}, \bibinfo{author}{He, Y.}, \& \bibinfo{author}{et~al.} (\bibinfo{year}{2023}).
\newblock \bibinfo{title}{Climax: A foundation model for weather and climate}.
\newblock {\it \bibinfo{journal}{arXiv preprint arXiv:2305.09996}\/}, .
\bibitem[{Pathak et~al.(2022)Pathak, Subramanian, Harrington \& et~al.}]{pathak2022fourcastnet}
\bibinfo{author}{Pathak, J.}, \bibinfo{author}{Subramanian, A.}, \bibinfo{author}{Harrington, M.}, \& \bibinfo{author}{et~al.} (\bibinfo{year}{2022}).
\newblock \bibinfo{title}{Fourcastnet: A global data-driven high-resolution weather model using adaptive fourier neural operators}.
\newblock {\it \bibinfo{journal}{arXiv preprint arXiv:2202.11214}\/}, .
\bibitem[{Qin et~al.(2017)Qin, Song \& Ma}]{qin2017dual}
\bibinfo{author}{Qin, L.}, \bibinfo{author}{Song, G.}, \& \bibinfo{author}{Ma, H.} (\bibinfo{year}{2017}).
\newblock \bibinfo{title}{A dual-stage attention-based recurrent neural network for time series prediction}.
\newblock {\it \bibinfo{journal}{IEEE Transactions on Neural Networks and Learning Systems}\/},  {\it \bibinfo{volume}{29}\/}, \bibinfo{pages}{4901--4914}.
\bibitem[{Raissi et~al.(2019)Raissi, Perdikaris \& Karniadakis}]{raissi2019pinns}
\bibinfo{author}{Raissi, M.}, \bibinfo{author}{Perdikaris, P.}, \& \bibinfo{author}{Karniadakis, G.~E.} (\bibinfo{year}{2019}).
\newblock \bibinfo{title}{Physics-informed neural networks: A deep learning framework for solving forward and inverse problems involving nonlinear partial differential equations}.
\newblock {\it \bibinfo{journal}{Journal of Computational Physics}\/},  {\it \bibinfo{volume}{378}\/}, \bibinfo{pages}{686--707}.
\bibitem[{Rasp et~al.(2020)Rasp, Pritchard \& Gentine}]{rasp2020weatherbench}
\bibinfo{author}{Rasp, S.}, \bibinfo{author}{Pritchard, M.}, \& \bibinfo{author}{Gentine, P.} (\bibinfo{year}{2020}).
\newblock \bibinfo{title}{Weatherbench: A benchmark dataset for data-driven weather forecasting}.
\newblock {\it \bibinfo{journal}{Quarterly Journal of the Royal Meteorological Society}\/},  {\it \bibinfo{volume}{146}\/}, \bibinfo{pages}{1747--1769}.
\bibitem[{Verma et~al.(2024)Verma, Heinonen \& Garg}]{verma2024climode}
\bibinfo{author}{Verma, Y.}, \bibinfo{author}{Heinonen, M.}, \& \bibinfo{author}{Garg, V.} (\bibinfo{year}{2024}).
\newblock \bibinfo{title}{Climode: Climate and weather forecasting with physics-informed neural odes}.
\newblock {\it \bibinfo{journal}{arXiv preprint arXiv:2404.10024}\/}, . \URLprefix \url{https://arxiv.org/abs/2404.10024}.
\newblock \bibinfo{note}{Accepted as ICLR 2024 Oral}.
\bibitem[{Yu et~al.(2017)Yu, Wu, Wang \& et~al.}]{yu2017spatiotemporal}
\bibinfo{author}{Yu, H.}, \bibinfo{author}{Wu, Z.}, \bibinfo{author}{Wang, S.}, \& \bibinfo{author}{et~al.} (\bibinfo{year}{2017}).
\newblock \bibinfo{title}{Spatiotemporal recurrent neural networks: A deep learning framework for traffic prediction}.
\newblock {\it \bibinfo{journal}{IEEE Transactions on Intelligent Transportation Systems}\/},  {\it \bibinfo{volume}{19}\/}, \bibinfo{pages}{2347--2356}.

\end{thebibliography}

\end{document}